\pdfoutput=1

\documentclass[11pt]{article}
\usepackage[ruled,vlined]{algorithm2e}
\usepackage{acl}

\usepackage{times}
\usepackage{latexsym}
\usepackage{graphicx}
\usepackage{amsmath}
\usepackage{stfloats}
\usepackage{amsfonts}
\usepackage[T1]{fontenc}

\usepackage[utf8]{inputenc}
\usepackage{booktabs}
\usepackage{microtype}

\title{A Simple and Effective Method to Improve Zero-Shot Cross-Lingual Transfer Learning}

\author{
Kunbo Ding\textsuperscript{\rm 1 \thanks{$^*$ Contribution during internship at Tencent Inc.}}, 
Weijie Liu\textsuperscript{\rm 1,2 \thanks{$^\dagger$ Corresponding author: Weijie Liu.}}, 
Yuejian Fang\textsuperscript{\rm 1}, 
Weiquan Mao\textsuperscript{\rm 2},
Zhe Zhao\textsuperscript{\rm 2}\\
\bf{Tao Zhu\textsuperscript{\rm 2},
Haoyan Liu\textsuperscript{\rm 2},
Rong Tian\textsuperscript{\rm 2},
Yiren Chen\textsuperscript{\rm 2}}\\
\textsuperscript{\rm 1}Peking University, Beijing, China  \textsuperscript{\rm 2}Tencent Research, Beijing, China\\
\footnotesize{kunbo\_ding@stu.pku.edu.cn, dataliu@pku.edu.cn, fangyj@ss.pku.edu.cn}\\
\footnotesize{\{weiquanmao, nlpzhezhao, mardozhu, haoyanliu, rometian, yirenchen\}@tencent.com}\\
}

\begin{document}
\maketitle
\begin{abstract}

Existing zero-shot cross-lingual transfer methods rely on parallel corpora or bilingual dictionaries, which are expensive and impractical for low-resource languages. To disengage from these dependencies, researchers have explored training multilingual models on English-only resources and transferring them to low-resource languages. However, its effect is limited by the gap between embedding clusters of different languages. To address this issue, we propose Embedding-Push, Attention-Pull, and Robust targets to transfer English embeddings to virtual multilingual embeddings without semantic loss, thereby improving cross-lingual transferability. Experimental results on mBERT and XLM-R demonstrate that our method significantly outperforms previous works on the zero-shot cross-lingual text classification task and can obtain a better multilingual alignment.

\end{abstract}

\section{Introduction}
In recent years, advances in multilingual models such as mBERT \citep{devlin2019bert}, XLM \citep{conneau2019cross}, XLM-R \citep{conneauetal2020unsupervised}, etc., after being fine-tuned with annotated data, have enabled significant improvements in many cross-lingual tasks. However, due to the lack of annotated data, some tasks in low-resource languages have not enjoyed this technological advancement. To solve this issue, the academic and industrial community began to focus on zero-shot cross-lingual transfer learning \citep{huang-etal-2019-unicoder, artetxe-etal-2020-cross}, which aims to fine-tune multilingual models with annotated data in high-resource languages and obtain a nice performance in low-resource language tasks. 

Some works aligned word embeddings between high- and low-resource languages through additional parallel sentence pairs \citep{artetxeschwenk2019massively,WeiW0XYL21,chi-etal-2021-infoxlm,pan-etal-2021-multilingual} or bilingual dictionaries \citep{Cao2020Multilingual,ijcai2020-533,Liu_Winata_Lin_Xu_Fung_2020}, so that high-resource fine-tuned models can be transferred to low-resource languages. Although this approach has achieved excellent results in many languages, parallel corpora and bilingual dictionaries are still prohibitively expensive, rendering it impracticable in some minority languages.

\begin{figure}[t]
	\centering
	\includegraphics[width=\linewidth]{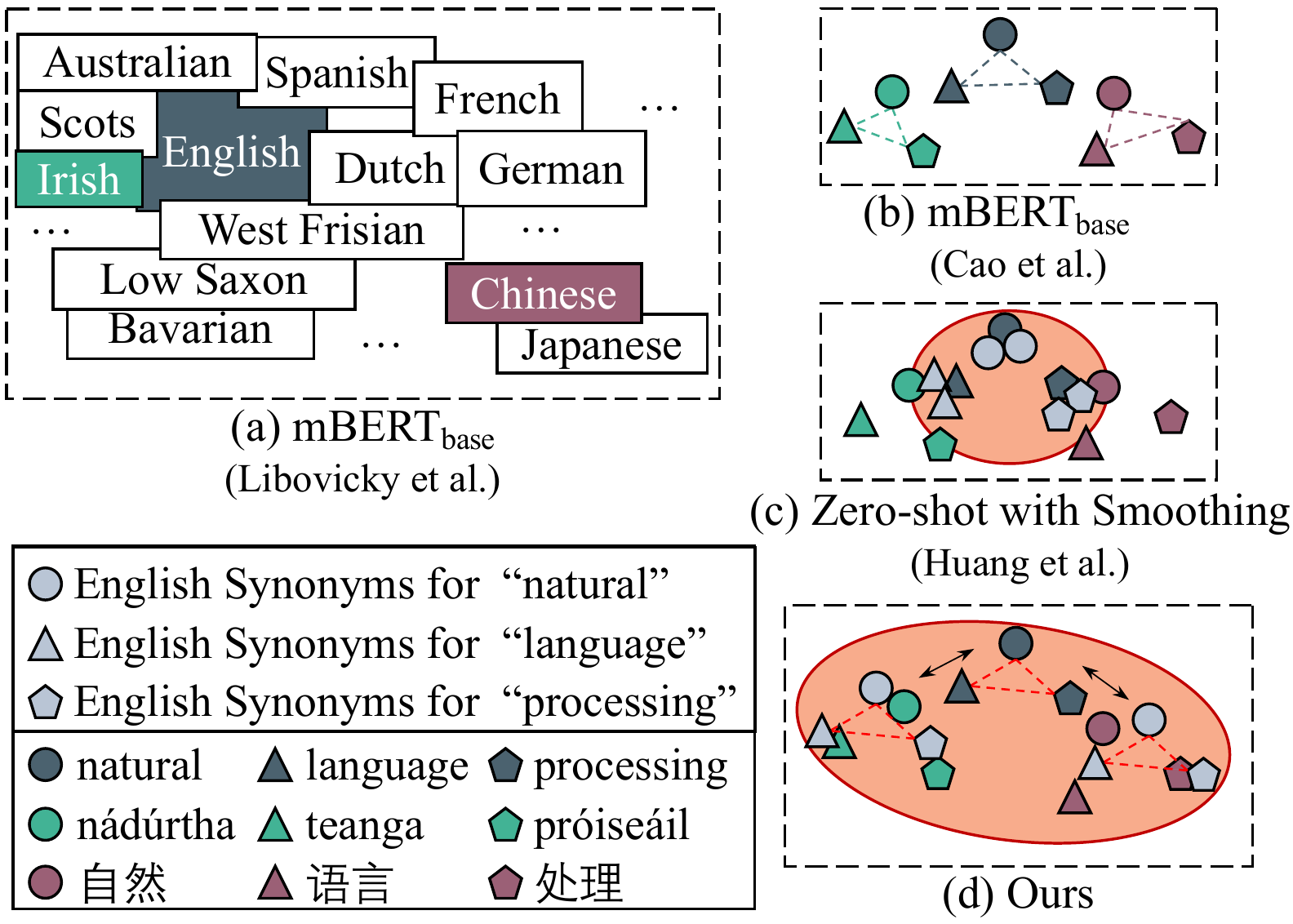}
	\caption{(a) Different languages clusters in mBERT. (b) The relative positions of "nature", "language" and "processing" are similar in English, Chinese and Irish \citep{Cao2020Multilingual}. (c) Using synonym augmentation to train a robust region covering words in other languages. (d) We align different languages and construct a suitable robust region by pushing the embeddings away and pulling the relative distance among words.}
	\label{fig:intro}
\end{figure}

To disengage from the dependence on parallel corpora or bilingual dictionaries \citep{wu-dredze-2019-beto,pmlr-v119-hu20b}, some studies have found that syntactic features in high-resource languages can improve zero-shot cross-lingual transfer learning \cite{meng-etal-2019-target,subburathinam-etal-2019-cross,ahmad-etal-2021-syntax,Ahmad_Peng_Chang_2021}. \citet{libovicky-etal-2020-language} found that the embeddings of different languages are clustered according to their language families, as shown in Figure \ref{fig:intro}a and \ref{fig:intro}b, which demonstrated that different languages are not aligned perfectly in mBERT \citep{Ameet-Mbert}. \citet{huang-etal-2021-improving-zero} tried adversarial training and randomized smoothing with English synonym augmentation to build robust regions for embeddings in the multilingual models, as illustrated in Figure \ref{fig:intro}c. In this way, models can output similar predictions for different language embeddings in the same robust region even they are not well aligned. However, the transferability of English synonym augmentation is limited because its robust region remains close to the English cluster, as shown in Figure \ref{fig:intro}c.

In this work, we select English as a high-resource language and follow the studies that do not require additional parallel corpora or bilingual dictionaries to improve cross-lingual transfer learning performance with minimal cost. For this purpose, three strategies are proposed to enlarge the robust region of English embeddings. The first strategy is called \textit{Embedding-Push}, which pushes the embedding of English to other language clusters. The second is \textit{Attention-Pull}, which constrains the relative position of the word embeddings to prevent the meaning from straying. The last strategy, named \textit{Robust target}, introduces a Virtual Multilingual Embedding (VME) to help the model build a suitable robust region, as shown in Figure \ref{fig:intro}d.

Experimental results on mBERT and XLM-R demonstrate that our method effectively improves the zero-shot cross-lingual transfer on classification tasks and outperforms a series of previous works. In addition, case studies show that our method improves the model through multilingual word alignment. Compared with existing works, our method has the following advantages. First, our method only needs English resources, which is suitable for low-resource languages. Second, our method can induce alignments in many languages without specifying the target language. Finally, our method is simple to implement and achieves effective experimental results. Our code
is publicly available\footnote{https://github.com/KB-Ding/EAR}.

\section{Method}
  \begin{figure}[t]
	\centering
	\includegraphics[width=\linewidth]{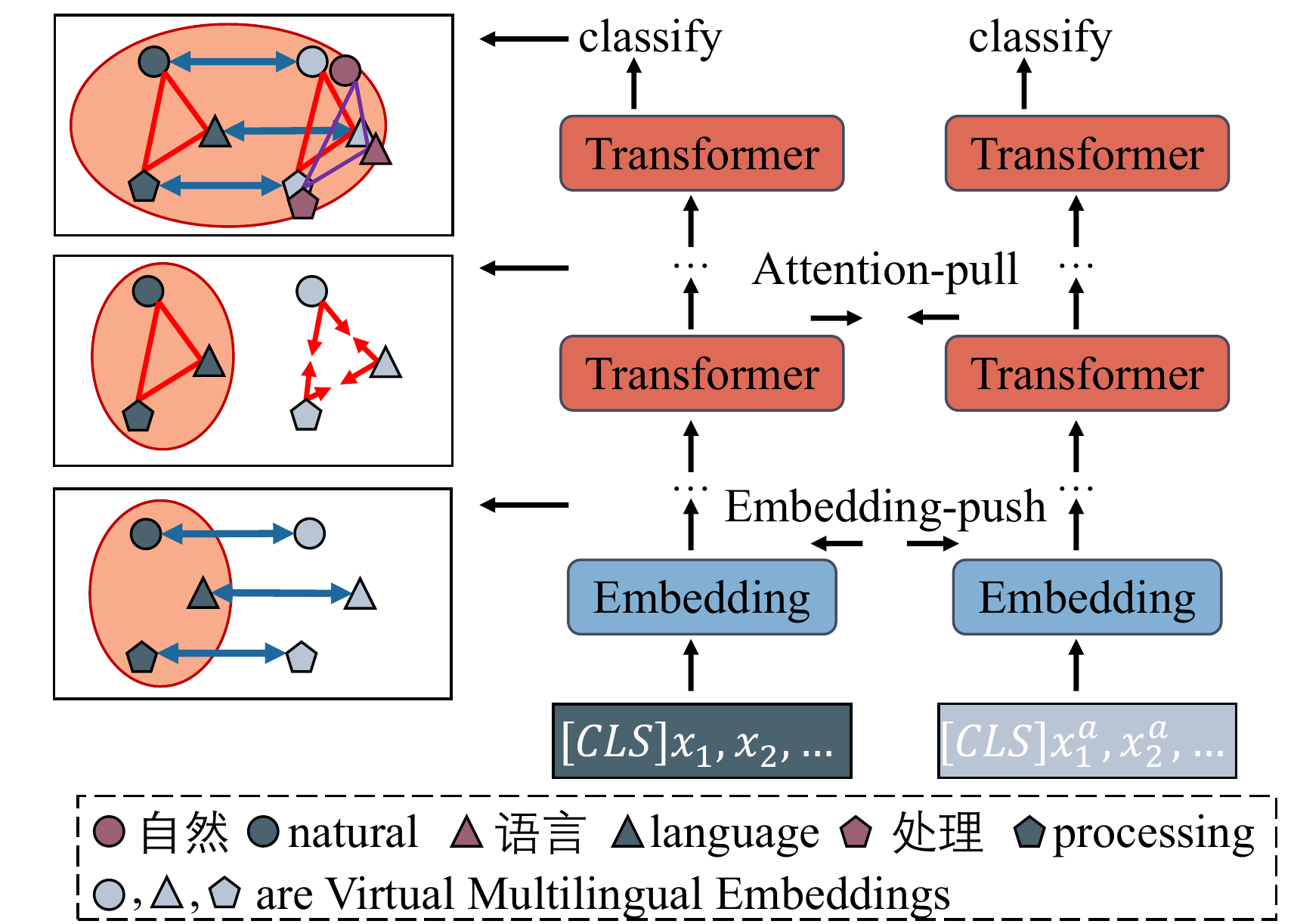}
	\caption{The two networks have tied weights. VMEs expand robust regions (orange circle) by aligning semantic-similar words in other languages. Note that VMEs do not specify the target language but improve multilingual performance, as shown in section \ref{sec: main}.}
	\label{fig:model}
\end{figure}

Given an English training batch $\mathcal{B}$, for a specific $\boldsymbol{x} \in \mathcal{B}$ consisting of words ($x_{1},x_{2},x_{3}$), we first follow \citet{huang-etal-2021-improving-zero} to generate an augmented example $\boldsymbol{x^{a}}=(x_{1}^{a},x_{2}^{a},x_{3}^{a})$ by randomly replacing $x_{i}$ with $x_{i}^{a}$ from the pre-defined English synonym set \citep{alzantot-etal-2018-generating}.
Then, we introduce three objective functions to get the Virtual Multilingual Embedding (VME) that provides a suitable robust region for zero-shot cross-lingual classification task as shown in Figure \ref{fig:model}. We describe the details in the following subsections.

\begin{table*}[ht!]
	\centering
	\scriptsize
	\def\arraystretch{1.0}%
	\setlength{\tabcolsep}{3.5pt}
	\begin{tabular}{@{}l c c c c c c c c c c c c c c c c}
	\toprule
	\textbf{Model}	&	\textbf{en}	&	\textbf{ar}	&	\textbf{bg}	&	\textbf{de}	&	\textbf{el}	&	\textbf{es}	&	\textbf{fr}	&	\textbf{hi}	&	\textbf{ru}	& \textbf{sw}   & \textbf{th}  &	\textbf{tr}	&	\textbf{ur}	&	\textbf{vi}	&	\textbf{zh}	  &   \textbf{\underline{avg.}}	\\ 
	\midrule
    mBERT$^\dagger$            & 80.8 & 64.3 & 68.0 & 70.0 & 65.3 & 73.5 & 73.4 & 58.9 & 67.8 & 49.7 & 54.1 & 60.9 & 57.2 & 69.3 & 67.8 & 65.4 	\\
	\quad +ADV$^\dagger$         & 81.9 & 64.9 & 68.3 & 71.7 & 66.5 & 74.4 & 74.5 & 59.6 & 68.8 & 48.8 & 50.6 & 61.7 & 59.2 & 70.0 & 69.4 & 66.0 \\
	 \quad +RS-RP$^\dagger$       & 82.6 & 65.4 & 68.7 & 70.5 & 67.2 & 75.0 & 74.1 & 59.8 & 69.5 & 48.4 & 50.5 & 59.7 & 57.9 & 70.5 & 69.7 & 66.0 \\
	 \quad +RS-DA$^\dagger$       & 81.0 & 66.4 & 69.9 & 71.8 & 68.0 & 74.7 & 74.2 & 62.7 & 70.6 & 51.1 & 55.7 & 62.9 & 60.9 & 71.8 & 71.4 & 67.6 \\
 	 \quad +Syntax$^\ddagger$ & 81.6 & 65.4	& 69.3 & 70.7	& 66.5	& 74.1	& 73.2 & 60.5 & 68.8 &	-	&	-	& 62.4	& 58.7 & 69.9 & 69.3	&  -  \\
	 
	 \quad + \textbf{Ours}       & \textbf{83.2} & \textbf{67.4} & \textbf{71.0} & \textbf{72.9} & \textbf{68.3} & \textbf{75.7} & \textbf{75.2} & \textbf{64.0} & \textbf{71.6} & \textbf{51.3} & \textbf{56.7} & \textbf{63.6} & \textbf{61.4} & \textbf{72.4} & \textbf{71.5} &  \textbf{\underline{68.4}} \\
	\midrule
    
    XLM-R$^*$  & 84.0 & 72.6 & \textbf{78.9} & 77.0 & 76.5 & 78.6 & 78.2 & 70.3 & 76.4 & 65.0 & 72.4 & 73.4 & 67.6 & 75.5 & 75.0 & 74.8 \\

    \quad +RS-DA$^*$      & 83.5 & 73.2 & 78.2 & 77.1 & 76.9 & 79.2 & 79.0 & 72.3 & \textbf{76.9} & 66.5 & 73.2 & 73.1 & 68.2 & \textbf{76.4} & 75.1 & 75.3 \\
	\quad + \textbf{Ours}  & \textbf{84.6} & \textbf{74.5} & 78.8 & \textbf{77.5} & \textbf{77.0} & \textbf{79.4} & \textbf{79.5} & \textbf{72.6} & 76.8 & \textbf{66.7} & \textbf{73.9} & \textbf{74.7} & \textbf{68.7} & \textbf{76.4} & \textbf{75.8} & \textbf{\underline{75.8}}  \\

	\bottomrule
	\end{tabular}
\caption{
Zero-shot cross-lingual transfer results on the XNLI. "$\dagger$" and "$\ddagger$" are taken from \citep{huang-etal-2021-improving-zero} and \citep{ahmad-etal-2021-syntax}, respectively. "$*$" is the result of our reimplementation.
}
\label{table:all_result_xnli} 
\end{table*}

\subsection{Embedding-push target}

The Embedding-Push target aims to make English embeddings leave their original cluster and robust region by pushing away ($\boldsymbol{x},\boldsymbol{x^{a}}$) in the embedding space. The pushed embedding can be viewed as the VME. The loss function is (\ref{eq:embedding}).
\begin{align}
   \label{eq:embedding}
       \ell _{EPT} &= - \frac{1}{|\mathcal{B}|} \sum_{\boldsymbol{x} \in \mathcal{B}} \left(M(E_{\boldsymbol{x}}) - M(E_{\boldsymbol{x^{a}}})\right)^2
\end{align}
where $E_{\boldsymbol{x}},E_{\boldsymbol{x^{a}}}$ denote the embedding output of $\boldsymbol{x}$ and $\boldsymbol{x^{a}}$, $M$ is the mean-pooling method. 
\subsection{Attention-pull target}
The self-attention matrices contain rich linguistic information \citep{clark-etal-2019-bert} and can be regarded as a 1-hop graph attention between the hidden states of words \citep{NIPS2017_3f5ee243, veličković2018graph}. The attention matrix represents the information transfer score between each pair of words, we regard it as the pulling force, so the attention matrix determines the relative linguistic positions of words in a sentence. We introduce the Attention-Pull target to encourage the relative linguistic position among ($x_{1}^{a},x_{2}^{a},x_{3}^{a}$) to be similar to ($x_{1},x_{2},x_{3}$) by fitting the middle layer multi-head attention matrices, as (\ref{eq:attention}).
\begin{align}
  \label{eq:attention}
  \ell _{APT} &= \frac{1}{|\mathcal{B}|H} \sum_{\boldsymbol{x} \in \mathcal{B}} \sum_{i}^{H} \left(A_{\boldsymbol{x}}^{i} -A_{\boldsymbol{x^{a}}}^{i}\right)^2
\end{align}
where $H$ is the number of attention head. Let $L$ denote the sequence length, $A^{i} \in \mathbb{R}^{L \times L}$ is the attention matrix corresponding to the i-th head. $\ell _{APT}$ alleviates the semantic loss of the VME.

\subsection{Robust target}
The robust target aims to build a robust region with the VME for the classification task. The hidden state of [CLS] in the last layer is taken to classify, as (\ref{eq:predict}). The model is trained by (\ref{eq:classify}).
\begin{align}
   \label{eq:predict}
    P_n &= \text{softmax}(\boldsymbol{W}h_{n}^{\text{[CLS]}}+\boldsymbol{b})\\
   \label{eq:classify}
   \ell _{CE} &= - \frac{1}{|\mathcal{B}|} \sum_{\boldsymbol{x} \in \mathcal{B}} \left(y\log P_{\boldsymbol{x}} + y\log P_{\boldsymbol{x^a}}\right)
\end{align}
where $\boldsymbol{W}$ and $\boldsymbol{b}$ are trainable parameters. $P_n$ is the prediction for $n$. $y$ denotes the gold label for each $\boldsymbol{x} \in \mathcal{B}$. The final training objective is to minimize three targets as (\ref{eq:total}):
\begin{align}
  \label{eq:total}
    \ell &= \ell _{CE}+\alpha\ell _{EPT}+\beta\ell _{APT}
\end{align}
where $\alpha$ and $\beta$ are hyperparameters.

\begin{table}[tb]
	\centering 
	\scriptsize
		\def\arraystretch{1.0}
	    \setlength{\tabcolsep}{3.5pt}
	    \begin{tabular}{l c c c c c c c c}
    \toprule
    \textbf{Model} & \textbf{en} & \textbf{de} & \textbf{es} & \textbf{fr} & \textbf{ja} & \textbf{ko} & \textbf{zh} & \textbf{\underline{avg.}} \\
   \midrule
    mBERT$^\dagger$  & 94.0 & 85.7 & 87.4 & 87.0 & 73.0 & 69.6 & 77.0 & 82.0 \\
    \quad +ADV$^\dagger$  & 93.7 & 86.5 & 88.5 & 87.8 & 76.1 & 75.3 & 80.4 & 84.0 \\
    \quad +RS-RP$^\dagger$  & \textbf{94.5} & 87.4 & 90.0 & 89.5 & 77.9 & 77.5 & 82.0 & 85.5 \\
    \quad +RS-DA$^\dagger$ & 93.5 & 87.8 & 88.8 & 88.8 & 79.3 & 78.3 & 81.5 & 85.4 \\
    \quad +Syntax$^\ddagger$	&	94.0 	&	85.9 &	89.1	&	88.2	&	75.8	&	76.3	& 80.7 & 84.3	\\
    \quad +\textbf{Ours}	&	94.2 	&	\textbf{87.9} &	\textbf{90.3}	&	\textbf{89.7}	&	\textbf{79.9}	&	\textbf{79.2}	& \textbf{82.4} & \textbf{\underline{86.2}}	\\
    \midrule
    XLM-R$^*$  & 94.4 & 88.9 & 89.8 & 89.2 & 78.2 & 78.4 & 81.4 & 85.7 \\
    \quad +RS-DA$^*$  & 94.7 & 88.8 & 89.7 & 90.0 & 78.7 & 80.2 & 82.3 & 86.3 \\
    \quad +\textbf{Ours}	&	\textbf{95.1} 	&	\textbf{89.0} &	\textbf{90.3}	&	\textbf{90.1}	&	\textbf{80.5}	&	\textbf{81.7}	& \textbf{83.1} & \textbf{\underline{87.1}}	\\
    \bottomrule
    \end{tabular}
	\caption{
	Experimental results on the PAWS-X across 7 languages. "$\dagger$" and "$\ddagger$" are taken from \citep{huang-etal-2021-improving-zero} and \citep{ahmad-etal-2021-syntax}, respectively. "$*$" is the result of our reimplementation.
	}
	\label{table:all_result_pawsx}
\end{table}

\section{Experiment}
\subsection{Dataset and setup}
\label{sec:setup}
We use $\text{mBERT}_{\text{base}}$ and $\text{XLM-R}_{\text{base}}$ to evaluate our method on XNLI \citep{conneau-etal-2018-xnli} and PAWS-X \citep{yang-etal-2019-paws} tasks, covering 17 languages. We consider English as the source language and other languages in test sets as low-resource target languages. More training details are in Appendix \ref{sec:details}. We set $\alpha$=1, $\beta$=0.1 and apply the Attention-Pull target at the 6-th layer. The analysis of hyperparameters is in Appendix \ref{sec:hyper}. We measure results with accuracy.

\subsection{Baseline methods }
For XLM-R, we consider \textbf{RS-DA} as a strong baseline because it achieves the best performance. For mBERT, we consider all the following baselines.

\textbf{Adv}: \citet{huang-etal-2021-improving-zero} uses adversarial training to build a robust region for cross-lingual transfer. They consider the most effective perturbation in each iteration.

\textbf{RS-RP}: \citet{huang-etal-2021-improving-zero} perturbs sentence embeddings with randomly sampled $\delta$ to smooth the classifier and build robust regions.

\textbf{RS-DA}: \citet{huang-etal-2021-improving-zero} augments training data with English synonym replacement to train a smooth classifier and build robust regions.

\textbf{Syntax}: \citet{ahmad-etal-2021-syntax} provides syntax features to mBERT by graph attention networks, which helps cross-lingual transfer. 

\subsection{Main results}
\label{sec: main}
As illustrated in Table \ref{table:all_result_xnli} and Table \ref{table:all_result_pawsx}. We can observe that: 1) Our method achieves up to 4.2\% and 1.4\% improvement on mBERT and XLM-R, respectively, outperforming existing works and demonstrating the effectiveness of our method. 2) Multiple low-resource languages benefit from our method. Based on mBERT, our method improves not only English-like languages such as \textbf{es} and \textbf{de} but also English-dissimilar \citep{littell-etal-2017-uriel} languages such as \textbf{tr} and \textbf{ko}. This result indicates that the VME we proposed helps align different languages in semantic space. 3) We avoid training each target language separately and achieves the best results in one epoch using the English-trained VME.

 \subsection{Ablation study}
 \label{sec:ablation}
\begin{table}[tb]
	\centering
	\scriptsize
		\def\arraystretch{0.9}
	    \setlength{\tabcolsep}{3.5pt}
	\begin{tabular}{l c c c c c c c c}
    \toprule
    \textbf{Model} &  \textbf{en}   &  \textbf{ar}   &  \textbf{bg}   &  \textbf{de}   &  \textbf{el}   &  \textbf{es}   &  \textbf{fr}   &  \textbf{hi} \\
    \midrule
    Ours & 83.2 & 67.4 & 71.0 & 72.9 & 68.3 & 75.7 & 75.2 & 64.0 \\
    w/o EPT & 82.8 & 67.0 & 71.2 & 72.7 & 67.6 & 75.5 & 75.1 & 63.4 \\
    w/o APT & 82.4 & 66.5 & 70.8 & 72.8 & 68.5 & 76.0 & 75.1 & 63.4 \\
    w/o both & 82.1 & 66.4 & 70.0 & 72.3 & 67.7 & 75.1 & 74.9 & 62.8 \\
    \midrule
     \textbf{Model}     &  \textbf{ru}   &  \textbf{sw}   &  \textbf{th}   &  \textbf{tr}   &  \textbf{ur}   &  \textbf{vi}   &  \textbf{zh}   &  \textbf{\underline{avg.}} \\
    \midrule
    Ours & 71.6 & 51.3 & 56.7 & 63.6 & 61.4 & 72.4 & 71.5  & \textbf{\underline{68.4}}  \\
    w/o EPT & 71.2 & 51.1 & 56.0 & 63.4 & 60.7 & 72.5 & 71.4 & \underline{68.1} \\
    w/o APT & 70.9 & 50.0 & 57.0 & 62.8 & 61.9 & 72.0 & 72.3 & \underline{68.2} \\
    w/o both     & 70.8 &48.3 & 54.6 & 61.0 & 61.0 & 71.5 & 71.5 & \underline{67.4} \\
    \bottomrule
    \end{tabular}
	\caption{Ablation experimental results of our method on the XNLI task. Experiments are based on mBERT. }
	\label{table:ablation}
\end{table}

As shown in Table \ref{table:ablation}, we perform ablation studies on Embedding-Push Target (\textbf{EPT}) and Attention-Pull Target (\textbf{APT}). We find that both EPT and APT are effective, but they can not perform well alone. Besides, removing the APT causes improvement in some languages, such as \textbf{zh} and \textbf{ur}. We attribute this to the fact that the EPT-guided VME is unstable without the APT, which improves performance in some languages but drops in more languages such as \textbf{en}, \textbf{ar}, \textbf{ru}, etc., resulting in poor average performance. Thus EPT and APT need to be combined for better performance.

\section{Analysis}

\begin{figure}[ht]
	\centering
	\includegraphics[width=0.95\linewidth]{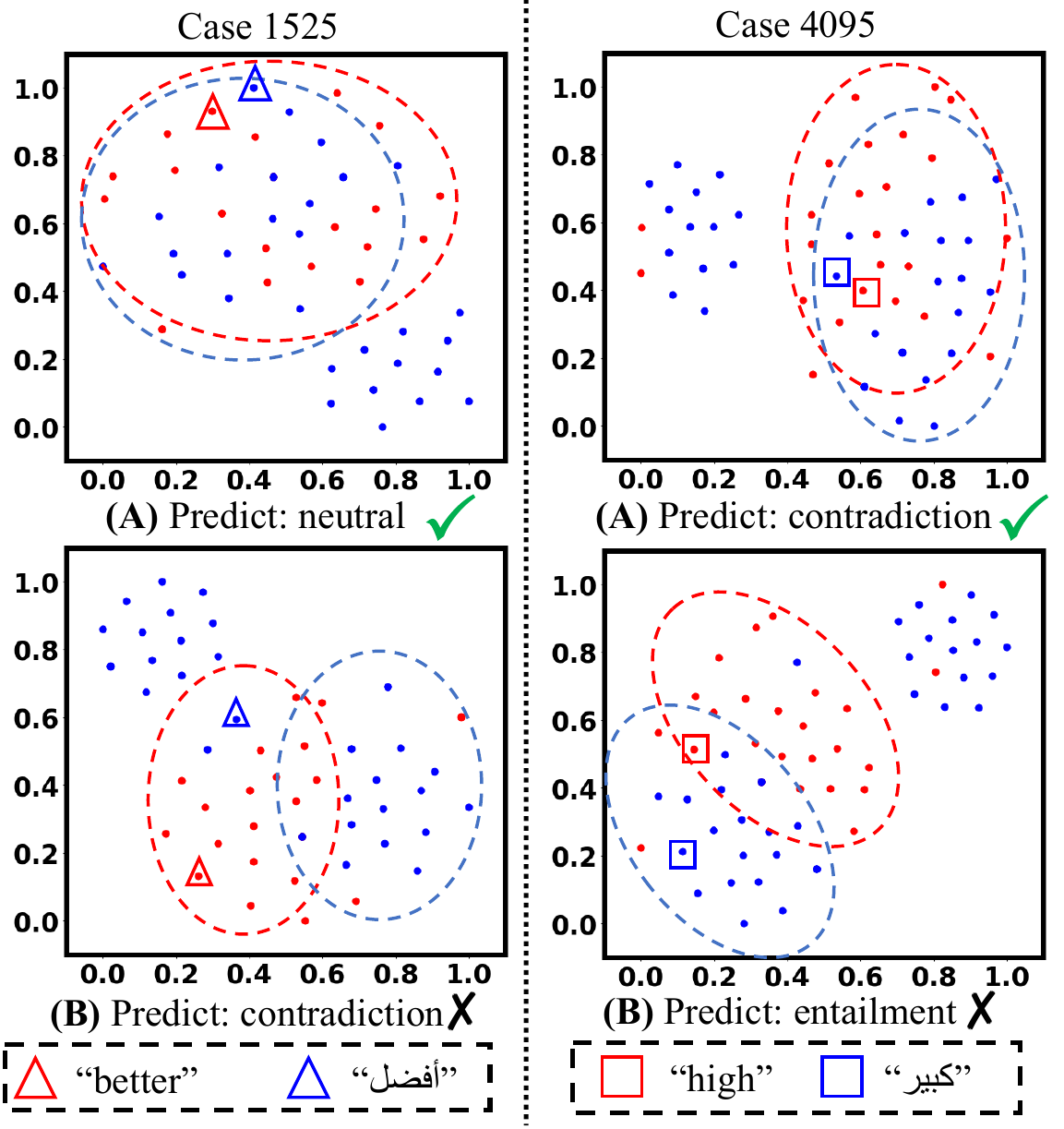}
	\caption{T-SNE visualization for word embeddings of English and Arabic translated sentences in XNLI test sets. Blue dots are Arabic words. Red dots are English words. (A) mBERT trained with our method. (B) mBERT trained with RS-DA.}
	\label{fig:Tsne}
\end{figure}

\begin{figure}[t]
	\centering
	\includegraphics[width=0.95\linewidth]{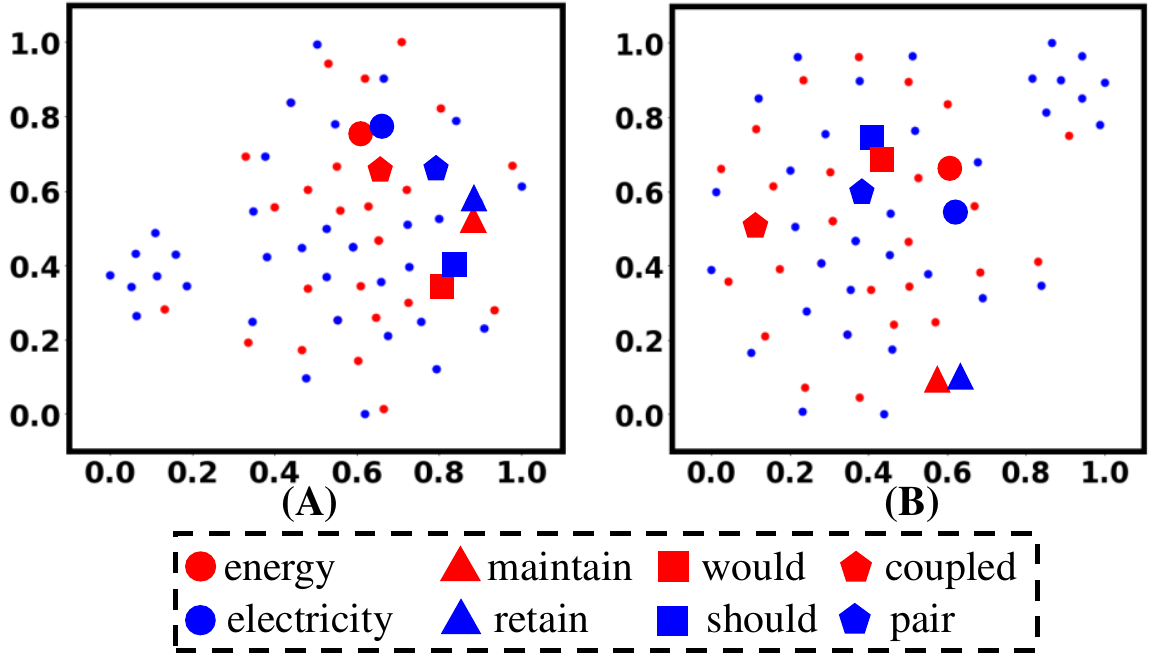}
	\caption{Visualization for English synonyms in the XNLI dataset using the embedding layer of mBERT. (A) Untrained. (B) Trained with our method.}
	\label{fig:emb}
\end{figure}

\subsection{Case study} 

To study the effects of VME, we do the T-SNE visualization for the word embeddings of parallel sentences, as shown in Figure \ref{fig:Tsne}. Compared with the RS-DA, our fine-tuned model aligns better across languages, and words are closer to their translations, leading to correct predictions. This observation shows that the VME can effectively help cross-lingual word alignment and improve the performance of the model. We choose Arabic for the case study because it can represent a class of languages far apart from English.

\begin{table}[!ht]
	\centering
	\scriptsize
	\def\arraystretch{0.9}
    \setlength{\tabcolsep}{3.5pt}
	\begin{tabular}{l c c c c c c c c}
    \toprule
    \textbf{Model} &  \textbf{en}   &  \textbf{es}   &  \textbf{de}   &  \textbf{fr}   &  \textbf{bg}   &  \textbf{ru}   &  \textbf{el}   &  \textbf{th} \\
    \midrule
    EPT + APT & 83.2 & 75.7 & 72.9 & 75.2 & 71.0 & 71.6 & 68.3 & 56.7 \\
    NT + APT & 82.7 & 75.6 & 72.5 & 75.2 & 70.6 & 71.0 & 67.9 & 55.9 \\
  EPT + SRPT & 83.1 & 76.0 & 73.2 & 74.9 & 70.7 & 71.4 & 68.9 & 56.5 \\
    \midrule
     \textbf{Model}     &  \textbf{sw}   &  \textbf{vi}   &  \textbf{ar}   &  \textbf{zh}   &  \textbf{hi}   &  \textbf{ur}   &  \textbf{tr}   &  \textbf{\underline{avg.}} \\
    \midrule
    EPT + APT & 51.3 & 72.4 & 67.4 & 71.5 & 64.0 & 61.4 & 63.6  & \textbf{\underline{68.4}}  \\
    NT + APT & 50.6 & 72.3 & 66.9 & 71.8 & 63.4 & 61.0 & 62.9 & \underline{68.0} \\
 EPT + SRPT  & 50.5 & 72.3 & 67.0 & 71.8 & 63.3 & 60.9 & 63.1 & \underline{68.2} \\
    \bottomrule
    \end{tabular}
	\caption{Results on the XNLI task when replacing some targets, based on the mBERT. We sort languages according to their differences from English \citep{littell-etal-2017-uriel}, from top left (small) to bottom right (big).}
	\label{table:mechanism}
\end{table}

\subsection{Effect of EPT}

To study the impact of EPT, we do the T-SNE visualization using the embedding layer of mBERT. As shown in Figure \ref{fig:emb}, some synonyms such as "coupled / pair" and "energy / electricity" are pushed away in the embedding layer trained with EPT, and some synonyms are still close to their original words. It indicates that the EPT push away synonyms selectively. We also try to replace the EPT in (\ref{eq:total}) with the Noise Target (NT), which perturbs word embeddings with Gaussian noise \citep{pmlr-v97-cohen19c}. As shown in Table \ref{table:mechanism}, we find that the EPT setting outperforms NT. One possible explanation could be that the noise in NT affects all English tokens and thus may hurt performance.

\begin{figure}[t]
	\centering
	\includegraphics[width=0.95\linewidth]{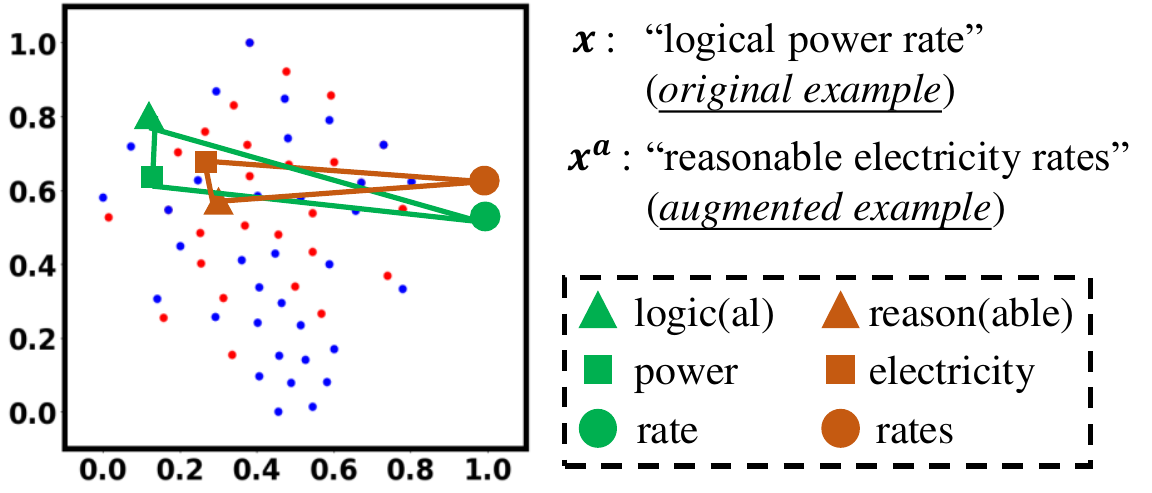}
	\caption{T-SNE visualization on the outputs of the mBERT trained with our method. The original words ($x$) and synonyms ($x^{a}$) are from the XNLI training sets.} 
	\label{fig:last-layer}
\end{figure}

\subsection{Effect of APT} 

To investigate the effects of APT, we replace the APT in (\ref{eq:total}) with the Sentence Representation Pull Target (SRPT). SRPT uses the mean squared error between sentence embeddings of $\boldsymbol{x}$ and $\boldsymbol{x^{a}}$ as the objective. Formally, $ \ell _{SRPT}=\frac{1}{|\mathcal{B}|} \sum_{i}^{|\mathcal{B}|}(\text{Sent}{(\boldsymbol{x})}-\text{Sent}(\boldsymbol{x^{a}}))^2 $, where Sent(x) represents the mean-pooled sentence embeddings \citep{reimers-gurevych-2019-sentence} obtained by the middle layer of the model. Results in Table \ref{table:mechanism} show that: 1) The average performance of SPRT is lower than that of APT. 2) The SRPT mainly improves performance on English-like languages, such as \textbf{es}, \textbf{de}, and \textbf{el}, while drops that of most English-dissimilar languages, such as \textbf{tr}, \textbf{hi}, \textbf{sw}, \textbf{ur}, etc. This phenomenon shows that SRPT suffers heavily from English training resources, biasing the VME towards English-like languages, which hurts the overall zero-shot cross-lingual transferability.

We perform T-SNE visualization on the outputs of the mBERT trained with our method. As shown in Figure \ref{fig:last-layer}, the synonym is still in the same relative position as the original word, which proves the effectiveness of APT.

\begin{table*}[!ht]
	\centering
	\scriptsize
	\def\arraystretch{1.0}
	\setlength{\tabcolsep}{3.5pt}
	\begin{tabular}{@{}l c c c c c c c c c c c c c c c c}
	\toprule
	\textbf{\textit{Source Language}}	&	\textbf{en}	&	\textbf{ar}	&	\textbf{bg}	&	\textbf{de}	&	\textbf{el}	&	\textbf{es}	&	\textbf{fr}	&	\textbf{hi}	&	\textbf{ru}	& \textbf{sw}   & \textbf{th}  &	\textbf{tr}	&	\textbf{ur}	&	\textbf{vi}	&	\textbf{zh}	  &   \textbf{\underline{avg.}}	\\ 
	\midrule
	\quad\quad \textbf{\textit{en}}   & \textbf{83.2} & 67.4 & 71.0 & 72.9 & 68.3 & 75.7 & \textbf{75.2} & 64.0 & 71.6 & \textbf{51.3} & 56.7 & 63.6 & 61.4 & 72.4 & 71.5 &  68.4 \\
    \quad\quad \textbf{\textit{de}}   & 79.6 &\textbf{68.7} 	&71.9 	&\textbf{77.7} 	&\textbf{68.8} 	&\textbf{76.2} 	&74.9 	&64.2 	&72.4 	&50.1 	&55.2 	&\textbf{64.0} 	&\textbf{62.8} 	&73.0 	&72.6 	&\textbf{\underline{68.8}} \\
   \quad\quad \textbf{\textit{ru}}   &78.5 	&68.2 	&\textbf{73.3} 	&73.1 	&\textbf{68.8}	&74.8 	&73.9 	&\textbf{65.8} 	&\textbf{75.7} 	&49.3 	&\textbf{57.2} 	&\textbf{64.0} 	&62.4 	&\textbf{73.4} 	&\textbf{73.7} 	&\textbf{\underline{68.8}} \\
	\bottomrule
	\end{tabular}
\caption{
Results of our method on the XNLI task when training mBERT with three source languages.
}
\label{table:source-languages} 
\end{table*}

\subsection{Effect of source language} 
In addition to \textbf{en}, both \textbf{de} and \textbf{ru} show preference as source languages in cross-lingual learning \citep{Iulia-revisiting}. We translate the training set into \textbf{de} and \textbf{ru} using OPUS-MT \citep{tiedemann-thottingal-2020-opus} models, as shown in Table \ref{table:source-languages}, the performance of our method can be further improved.

\begin{table}[t]
	\centering 
	\scriptsize
 	\setlength{\tabcolsep}{5mm}{
    \begin{tabular}{l c c}
    \toprule
    \textbf{scale} & \textbf{size of dictionary} & \textbf{\underline{XNLI result}} \\
   \midrule
    1.0  & 49975 & \textbf{68.424}\\
    0.75  & 37481 & \textbf{68.392}\\
    0.5  & 24987 & \textbf{68.218}\\
    0.25  & 12493 & \textbf{68.080}\\
    \bottomrule
    \end{tabular}
    }
	\caption{
 Results on the XNLI task when using the scaled synonym dictionaries for data augmentation.
	}
	\label{table:dictionary}
\end{table}

\subsection{Effect of dictionary size} 

\begin{figure}[t]
	\centering
	\includegraphics[width=0.95\linewidth]{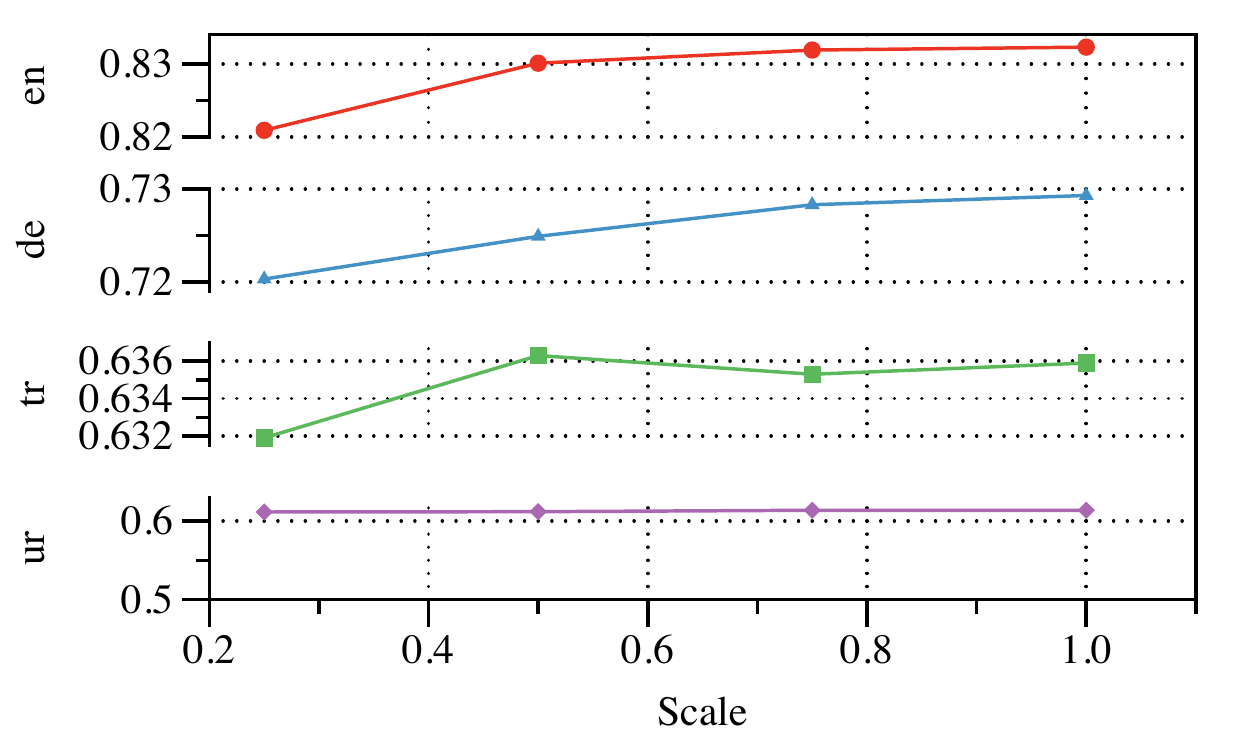}
	\caption{Results on XNLI test sets of four languages when using scaled synonym dictionaries in our method.}
	\label{fig:dic_size}
\end{figure}

The data augmentation in our method relies on the size of pre-defined synonym dictionary. As shown in Table \ref{table:dictionary} and Figure \ref{fig:dic_size}, we can observe that: 1) The overall performance decreases as the dictionary size decreases. 2) Some languages are not sensitive to the dictionary size, such as \textbf{tr} and \textbf{ur}. 3) The performance of \textbf{en}, \textbf{de}, and \textbf{tr} degrades significantly when the dictionary size is scaled from 0.5 to 0.25. This phenomenon may be related to some important synonyms in the dictionary, which are effective for cross-lingual transfer learning.

\section{Conclusion}
To get rid of the dependence on parallel corpora, enable cross-lingual transfer to low-resource languages, we propose Embedding-Push, Attention-Pull, and Robust targets to combat the influence of language clusters in multilingual models. Experimental results demonstrate that our method outperforms previous works and obtains better-aligned embeddings when trained with only English.

\bibliography{anthology,custom}
\bibliographystyle{acl_natbib}

\appendix
\setcounter{table}{0}   
\setcounter{figure}{0}
\renewcommand{\thetable}{A.\arabic{table}}
\renewcommand{\thefigure}{A.\arabic{figure}}

\begin{table*}[ht!]
    \centering
    \scriptsize
    \def\arraystretch{1.2}%
    \setlength{\tabcolsep}{3.5pt}
    \begin{tabular}{l|ccccccccccccccc|r} 
    \toprule
    \textbf{Layer} & \textbf{en} & \textbf{ar} & \textbf{bg} & \textbf{de} & \textbf{el} & \textbf{es} & \textbf{fr} & \textbf{hi} & \textbf{ru} & \textbf{sw} & \textbf{th} & \textbf{tr} & \textbf{ur} & \textbf{vi} & \textbf{zh} & \underline{\textbf{avg.}}            \\ 
    \midrule
3          & 83.23 & 67.17 & 71.44 & 73.33 & 68.06 & 75.99 & 74.89 & 63.27 & 70.94 & 51.00 & 56.61 & 63.23 & 61.04 & 72.30 & 71.66 & \underline{68.28}           \\
6 & 83.05 & 67.01 & 70.88 & 72.63 & 67.98 & 76.05 & 74.91 & 62.99 & 71.82 & 51.28 & 56.81 & 63.53 & 61.44 & 72.48 & 71.48 & \underline{\textbf{68.29}}  \\
9          & 82.87 & 67.56 & 71.22 & 73.05 & 68.36 & 75.81 & 74.63 & 63.65 & 71.14 & 50.96 & 56.75 & 62.97 & 61.00 & 72.55 & 71.68 & \underline{68.28}           \\
12         & 83.05 & 67.05 & 70.56 & 72.81 & 68.22 & 75.55 & 75.35 & 63.35 & 71.48 & 50.82 & 56.71 & 63.11 & 60.30 & 72.48 & 71.68 & \underline{68.17}           \\
    \bottomrule
    \end{tabular}
\caption{
Results of the XNLI task when we apply the Attention-Pull target at different layers of mBERT.
}
\label{table:hpyer_layer}
\end{table*}

\begin{table*}[ht!]
    \centering
    \scriptsize
    \def\arraystretch{1.2}%
    \setlength{\tabcolsep}{3.5pt}
    \begin{tabular}{l|ccccccccccccccc|r} 
    \toprule
    \textbf{\text{$\beta$}} & \textbf{en} & \textbf{ar} & \textbf{bg} & \textbf{de} & \textbf{el} & \textbf{es} & \textbf{fr} & \textbf{hi} & \textbf{ru} & \textbf{sw} & \textbf{th} & \textbf{tr} & \textbf{ur} & \textbf{vi} & \textbf{zh} & \textbf{\underline{avg.}}            \\ 
    \midrule
   0.1           & 83.23       & 67.41       & 71.04       & 72.93       & 68.28       & 75.75       & 75.19       & 63.99       & 71.64       & 51.28       & 56.73       & 63.59       & 61.38       & 72.44       & 71.50       & \underline{\textbf{68.42}}  \\
0.2           & 83.15       & 67.05       & 71.38       & 73.79       & 68.34       & 75.99       & 75.01       & 63.53       & 71.58       & 50       & 56.41       & 63.21       & 60.52       & 72.40       & 71.88       & \underline{68.32}           \\
0.3           & 83.09       & 67.47       & 71.52       & 72.99       & 68.44       & 75.65       & 75.03       & 63.57       & 71.42       & 50.76       & 55.87       & 63.29       & 61.26       & 72.63       & 71.58       & \underline{68.30}           \\
0.5           & 83.01       & 67.15       & 70.58       & 72.95       & 68.10       & 75.87       & 74.99       & 62.99       & 71.64       & 50.34       & 56.37       & 63.61       & 60.82       & 72.57       & 72.18       & \underline{68.21}           \\
0.7           & 82.69       & 66.83       & 71.08       & 72.87       & 68.14       & 75.89       & 74.43       & 63.15       & 71.00       & 51.60       & 56.57       & 63.15       & 60.88       & 72.16       & 71.82       & \underline{68.15}           \\
0.9           & 82.51       & 66.83       & 71.00       & 72.87       & 68.50       & 75.65       & 75.01       & 62.99       & 71.30       & 50.68       & 55.77       & 63.29       & 61.38       & 72.42       & 71.54       & \underline{68.12}           \\
    \bottomrule
    \end{tabular}
\caption{
The experimental results of the XNLI task based on mBERT when $\beta$ takes different values, where $\alpha$=1.
}
\label{table:hpyer_beta}
\end{table*}

\begin{table*}[ht!]
    \centering
    \scriptsize
    \def\arraystretch{1.2}%
    \setlength{\tabcolsep}{3.5pt}
    \begin{tabular}{l|ccccccccccccccc|r} 
    \toprule
    \textbf{$\alpha$} & \textbf{en} & \textbf{ar} & \textbf{bg} & \textbf{de} & \textbf{el} & \textbf{es} & \textbf{fr} & \textbf{hi} & \textbf{ru} & \textbf{sw} & \textbf{th} & \textbf{tr} & \textbf{ur} & \textbf{vi} & \textbf{zh} & \underline{\textbf{avg.}}            \\ 
    \midrule
0.6            & 82.85       & 67.35       & 71.64       & 73.03       & 68.32       & 75.65       & 74.57       & 63.37       & 71.56       & 50.56       & 56.39       & 63.67       & 61.16       & 72.44       & 72.02       & \underline{68.30}           \\
0.8            & 82.87       & 67.23       & 70.96       & 73.17       & 68.68       & 75.23       & 74.87       & 63.53       & 71.26       & 50.86       & 56.45       & 63.21       & 61.30       & 72.55       & 72.02       & \underline{68.28}           \\
1              & 83.23       & 67.41       & 71.04       & 72.93       & 68.28       & 75.75       & 75.19       & 63.99       & 71.64       & 51.28       & 56.73       & 63.59       & 61.38       & 72.44       & 71.50       & \underline{\textbf{68.42}}  \\
1.2            & 83.19       & 67.03       & 71.08       & 72.97       & 67.86       & 75.87       & 74.75       & 63.49       & 71.50       & 51.60       & 56.37       & 63.21       & 60.88       & 72.59       & 71.28       & \underline{68.24}           \\
1.4            & 83.09       & 67.03       & 71.44       & 73.35       & 68.78       & 75.79       & 74.51       & 63.23       & 71.50       & 51.14       & 56.35       & 63.45       & 60.72       & 72.75       & 71.60       & \underline{68.32}           \\
1.6            & 83.19       & 67.05       & 71.50       & 73.23       & 68.18       & 76.25       & 74.77       & 63.43       & 71.06       & 51.10       & 56.43       & 62.95       & 60.52       & 72.38       & 71.98       & \underline{68.27}           \\
1.8            & 83.29       & 67.09       & 71.44       & 73.51       & 68.54       & 75.85       & 74.79       & 63.83       & 71.36       & 50.78       & 56.47       & 63.23       & 60.86       & 72.59       & 71.98       & \underline{68.37}           \\
    \bottomrule
    \end{tabular}
\caption{
The experimental results of the XNLI task based on mBERT when $\alpha$ takes different values, where $\beta$=0.1.
}
\label{table:hpyer_alpha}
\end{table*}

\section{Implementation details}
\label{sec:details}
\paragraph{Dataset}
XNLI is the cross-lingual natural language inference task. PAWS-X is used to determine whether two sentences paraphrase each other. The augmentation datasets are obtained from \citet{huang-etal-2021-improving-zero}. They augmented 3 and 10 examples for each sentence in XNLI and PAWS-X by synonym replacement, respectively. The pre-defined English synonym set is from \citet{alzantot-etal-2018-generating}. The scripts for splitting training, test, and validation sets are provided by XTREME \citep{pmlr-v119-hu20b}.

\paragraph{Setup}
The $\text{mBERT}_\text{base}$ and $\text{XLM-R}_\text{base}$ are obtained from Huggingface's \textit{transformers} package \citep{wolf-etal-2020-transformers}. The maximum sequence length is set as 128. The learning rate is set as 2e-5. Our method is trained for one epoch with the batch size of 32. other models are trained following \citet{pmlr-v119-hu20b} and \citet{huang-etal-2021-improving-zero}.

\paragraph{Input construction}
Both XNLI and PAWS-X are sentence pair classification tasks. Taking mBERT as an example, for each $\text{s}_{\text{1}}$, $\text{s}_{2}$ and augmented $\text{s}_{1}^{\text{a}}$, $\text{s}_{2}^\text{a}$ in the training data, we set $\boldsymbol{x}$ as [CLS]$\text{s}_{1}$[SEP]$\text{s}_2$[SEP], $\boldsymbol{x^a}$ as [CLS]$\text{s}_{1}^{\text{a}}$[SEP]$\text{s}_{2}^\text{a}$[SEP]. Then, we take $\boldsymbol{x}$ and $\boldsymbol{x^a}$ as the input of our method in Figure \ref{fig:model}, [CLS] token is used for classification.

\section{Hyperparameter analysis}
\label{sec:hyper}
There are three main hyperparameters in our method that need to be adjusted. 1) We need to determine which layer is most effective for applying Attention-Pull target. 2) We need to determine the weight of $\beta$ in the final loss. 3) We need to determine the weight of $\alpha$ in the final loss. We conduct experiments on XNLI task based on mBERT.

For 1), we first set $\alpha$=1 and $\beta$=1, then apply the Attention-Pull target on the \{3, 6, 9, 12\} layers respectively, and the results are shown in Table \ref{table:hpyer_layer}. We find that applying the Attention-Pull target to all layers works well. The most significant improvement is achieved at the 6-th layer and the minimal improvement is achieved at the last layer, which may be related to the quality of sentence representation at different layers of the model \citep{carlsson2021semantic, merchant-etal-2020-happens}.

For 2), we apply the Attention-Pull target at the 6-th layer and set $\alpha$=1, then select $\beta$ from \{0.1, 0.2, 0.3, 0.5, 0.7, 0.9\}. The experimental results are shown in Table \ref{table:hpyer_beta}. First, we find that model performance improved when using any of the above $\beta$ values. Second, we also find that the improvement becomes significant as $\beta$ decreases, we attribute this phenomenon to the fact that the Attention-Pull target should not over-focus on features of the English corpus but should help the VME capture features in other language clusters. Note that this result does not mean that the Attention-Pull target is unnecessary, as ablation experiments in section \ref{sec:ablation} show that the Attention-Pull target can improve the model. Finally, the best experimental result is obtained when $\beta$=0.1.

For 3), we apply the Attention-Pull target at the 6-th layer and set $\beta$=0.1, then select $\alpha$ from \{0.6, 0.8, 1.0, 1.2, 1.4, 1.6, 1.8\}. Results are shown as Table \ref{table:hpyer_alpha}. We find that the best performance is achieved when $\alpha$ is 1.0. The performance is also improved when using other $\alpha$ values, which shows that the Embedding-Push target can robustly improve the cross-lingual transferability of models. Therefore, in our main experiments, we set $\alpha$=1.0, $\beta$=0.1 and apply the Attention-Pull target at the 6-th layer.

\section{Analysis on XLM-R}
We perform analysis based on XLM-R, the results are shown in Table \ref{table:XLMR-analysis}.

\begin{table}[tb]
	\centering
	\scriptsize
	\def\arraystretch{1.0}
    \setlength{\tabcolsep}{4.5pt}
	\begin{tabular}{l c c c c c c c c}
    \toprule
    \textbf{Model} &  \textbf{en}   &  \textbf{es}   &  \textbf{de}   &  \textbf{fr}   &  \textbf{bg}   &  \textbf{ru}   &  \textbf{el}   &  \textbf{th} \\
    \midrule
    EPT + APT & 84.6 & 79.4 & 77.5 & 79.5 & 78.8 & 76.8 & 77.0 & 73.9 \\
    NT + APT & 84.4 & 79.4 & 77.2 & 79.0 & 78.8 & 76.7 & 76.4 & 73.4 \\
  EPT + SRPT & 84.4 & 80.0 & 77.8 & 79.2 & 78.5 & 76.8 & 77.1 & 74.2 \\
    \midrule
     \textbf{Model}     &  \textbf{sw}   &  \textbf{vi}   &  \textbf{ar}   &  \textbf{zh}   &  \textbf{hi}   &  \textbf{ur}   &  \textbf{tr}   &  \textbf{\underline{avg.}} \\
    \midrule
    EPT + APT & 66.7 & 76.4 & 74.5 & 75.8 & 72.6 & 68.7 & 74.7  & \textbf{\underline{75.8}}  \\
    NT + APT & 67.3 & 76.3 & 73.6 & 75.2 & 72.2 & 67.7 & 74.1 & \underline{75.5} \\
 EPT + SRPT  & 65.2 & 76.6 & 73.5 & 75.8 & 72.5 & 68.7 & 74.4 & \underline{75.6} \\
    \bottomrule
    \end{tabular}
	\caption{Results on the XNLI task when replacing some targets, based on the XLM-R.}
	\label{table:XLMR-analysis}
\end{table}

\end{document}